# Real-centric Consistency Learning for Deepfake Detection

Ruiqi Zha, Zhichao Lian, Qianmu Li, Siqi Gu
Nanjing University of Science and Technology, China

## Abstract

*Most of previous deepfake detection researches bent their efforts to describe and discriminate artifacts in human perceptible ways, which leave a bias in the learned networks of ignoring some critical invariance features intra-class and underperforming the robustness of internet interference. Essentially, the target of deepfake detection problem is to represent natural faces and forgery faces at the representation space discriminatively, and it reminds us whether we could optimize the feature extraction procedure at the representation space through constrain-ing intra-class consistence and inter-class inconsistence to bring the intra-class representations closer and push the inter-class representations apart. Therefore, inspired by contrastive representation learning, we tackle the deepfake detection problem through learning the invariant repre-sentations of both classes and propose a novel real-centric consistency learning method. We constraint the repre-sentation from both the sample level and the feature level. At the sample level, we take the procedure of deepfake synthesis into consideration and propose a novel forgery semantical guided pairing strategy to mine latent generation-related features. At the feature level, based on the center of natural faces at the representation space, we design a hard positive mining and synthesizing method to simulate the potential marginal features. Besides, a hard negative fusion method is designed to improve the discrimination of negative marginal features with the help of supervised contrastive margin loss we developed. The effectiveness and robustness of the proposed method have been demonstrated through extensive experiments.*

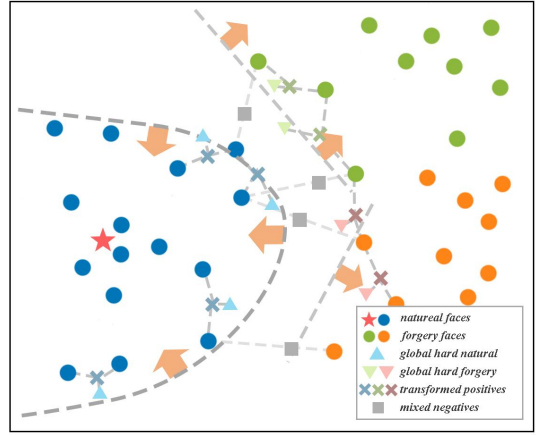

Figure 1: In this paper, we tackle the deepfake detection problem through learning the invariant representations of both classes through narrowing the intra-class discrepancy and widening the inter-class gap at the sample level and the feature level, instead of defining and representing artifacts at a specific domain in human perceptible ways.

## 1. Introduction

Deepfake is a type of fake images or videos, manipulated by the deep learning methods. Deepfake methods develop gradually as the gradual up growth of the deep learning methods [38, 39, 40]. Those methods could provide indistinct deepfake images or videos including subtle forgery clues. To prevent the social hazard aroused by the deepfake methods, deepfake detection has become an attractive research topic.

The core entry point of deepfake detection is to find out the artifacts manipulated by generative methods. Those artifacts could be roughly divided into two categories: visual biological artifacts [6, 12, 15] and facial semantical irrelevant artifacts [3, 4, 5, 7, 8, 9, 10, 11, 13, 14, 22]. Early works extracted artifacts through handcraft features to discover the forgery patterns [7, 8, 9, 10] or used shallow networks to extract low-level manipulated clues [3, 4, 5]. With the prosperous development of deep learning, more deep neural network frameworks [16, 17, 18, 19] with strong feature extraction ability have been proposed, which help recent works extract more precise facial landmarks to find out the unnatural micro-expressions [12] or transform the deepfake detection problem into other fields, like multi-task learning [22], fine-grained classification [13], noise detection [14], etc.

Many previous works have shown their effectiveness on the large-scale open-source face forgery datasets [1, 2]. However, in a more realistic condition, videos could be compressed for their transmission through the internet, which might add interference on the natural videos or erase artifacts on the fake videos either. As the mainstream generative methods are relatively fixed for a period of time, the robustness against compression is a more important property for detection methods in actual scenarios. However, most of the existing detection methods are based on discriminative models and extract artifacts from the fake aspect, which make a bias existence in the learned network about the discrepancy between fake regions and natural regions while ignoring the consistency inner that discriminates natural faces and forgery faces essentially. Therefore, the erased artifacts caused by compression is potentially catastrophic, and let alone the invalidation of facial landmarks.

Essentially, the target of deepfake detection problem is to represent natural faces and forgery faces at the representation space discriminatively, and it reminds us whether we could optimize the feature extraction procedure at the representation space through constraining intra-class consistence and inter-class inconsistence to bring the intra-class representations close and push the inter-class representations apart. Inspired by the excellent performance of contrastive learning achieved at the representation learning field [23, 24, 25, 26, 27, 28], instead of defining and representing artifacts at specific domain, we locate the deepfake detection problem on learning the invariant representations for natural faces and forgery faces both.

Similar to the human cognitive processes, contrastive learning methods learn from multi-views of an identical object to keep the consistence of views from the same object in the representation space and punish the views sharing high degree of similarity which belong to different objects through maximizing the lower bound of mutual information between views. Intuitively, the multi-views construction strategy is critical for contrastive learning methods, and the common way is augmenting twice of identical image as two views of an object subject under the setup of unsupervised learning. However, the latent relationships within data and how to utilize those relationships to guide a more effectiveness representation learning are rare discussed because of lacking label information. In addition, the improvement of positives construction strategy remains stay at the sample level which influences the network updates indirectly and does not independent from human intuition. A solution is to make constraints at the representation space, narrowing the intra-class discrepancy and widening the inter-class gap like restraining the distance between features or simulating the potential marginal features.

Based on the analysis above, we propose a novel real-centric consistency learning method for deepfake detection which first introduce contrastive learning into deepfake detection field. Specifically, we constraint feature extraction from two aspects, namely sample level and feature level. At the sample level, to mine latent generation-related features, we take the procedure of deepfake synthesis into consideration and propose a novel forgery semantical guided pairing strategy. At the feature level, based on the centers of natural faces at the representation space, we design a hard positive mining and synthesizing method to simulate the potential marginal features. In addition, we design a hard negative fusion method, as an auxiliary, where a supervised contrastive loss is developed with margin to improve the discrimination of marginal features.

Our contributions could be summarized as follows:
- We tackle the deepfake detection problem through learning the invariant representations of both classes instead of representing artifacts only in human perceptible ways and introduce contrastive learning into deepfake detection field first as far as we know.
- At sample level, a forgery semantical guided pairing strategy combining image information and temporal information as well as semantical information is proposed to mine latent generation-related features and improve the effectiveness of feature extraction.
- At feature level, a real-centric hard feature fusion method is designed to mine and synthesis hard features based on the center of natural faces at the representation space. As an auxiliary, the supervised contrastive loss is developed with margin to narrow the discrepancy of natural faces at the representation space and improve the discrimination of marginal features.

## 2. Related work

**Deepfake Detection.** Following the core idea of finding out the manipulation clues, existing deepfake detection methods could be divided into two categories: visual biological artifacts based methods [6, 12, 15] and facial semantical irrelevant artifacts based methods [3, 4, 5, 7, 8, 9, 10, 11, 13, 14, 22]. Early works extracted handcraft features to discover the forgery patterns [7, 8, 9, 10] or used a shallow network to extract low-level features [3, 4, 5]. With the prosperity of deep learning, generative methods could manipulate more lifelike forgery faces with subtle visual clues which early methods were hard to deal with, in the meantime, more well performed detection methods were proposed [12, 15, 11, 13, 14].

Recent visual biological artifacts based methods focus on unnatural micro-expressions [12] or more complex facial semantical information [15]. Researchers in [12] detected deepfake videos through temporal modeling on precise geometric features, and they calibrated each facial landmarks within frames to enhance the precision of facial geometric features and constructed a two-stream RNN for sufficient exploitation of temporal features. Targeting at high-level semantic irregularities in mouth movements, Haliassos et al. [15] used a pertained frozen feature extractor to obtain embedding features sensitive to mouth movements. After that, a pertained classifier was fine-tuned to detect fake videos based on semantically high-level irregularities in mouth motion. Although some of recent proposed visual biological artifacts based methods obtain performances robust against compression, the sensibility of facial landmarks and the requirement of temporal inputs make them hardly perform well under the condition of inconspicuous motion of facial landmarks or the discontinuity of frames (e.g. frame rate decreases or frame loss) which were commonly caused by Internet compression.

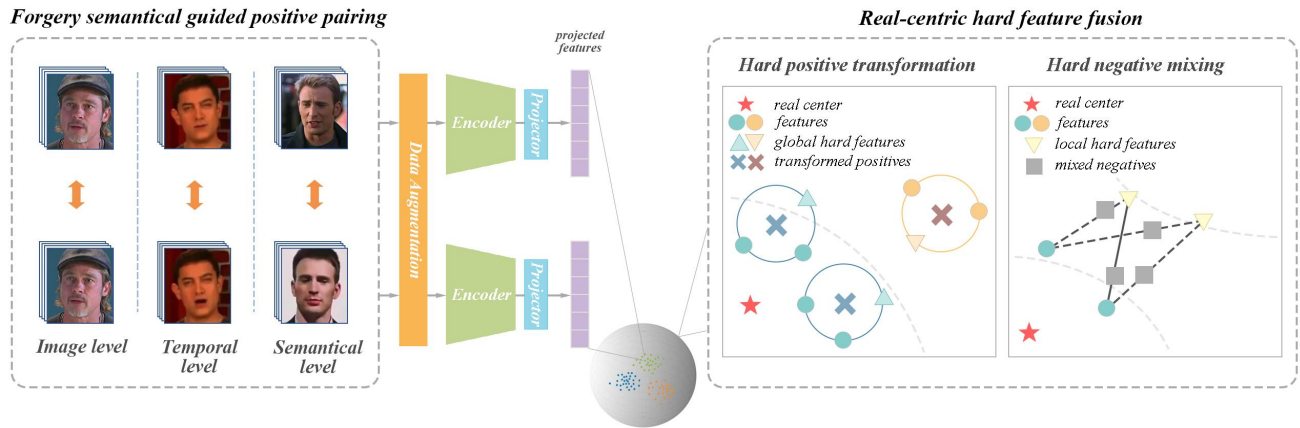

Figure 2: Instead of defining and representing artifacts at specific domain, we optimize the feature extraction procedure at the representation space through constraining intra-class consistence and inter-class inconsistence. At the sample level, we propose a novel forgery semantical guided pairing strategy to mine latent generation-related features. At the feature level, based on the center of the natural face, we design a hard positive mining and synthesizing method to simulate the potential marginal features, and a hard negative fusion method is proposed to improve the discrimination of marginal features with the help of supervised contrastive margin loss we developed.

Previous researches revealed that subtle forgery artifacts or compression errors could be well described in the frequency domain [11, 14, 21], and thus many recent facial semantical irrelevant artifacts based works detected deepfakes using frequency features. Due to that the conventional frequency domain do not match the shift-invariance and local consistency, being infeasible to vanilla CNN, [11] proposed frequency-aware decomposition and local frequency statistics to deeply mine the frequency-aware clues, and to learn forgery patterns through a cross-attention powered two-stream networks. Luo et al. [14] observed that image noises removed color textures and exposed discrepancies between authentic and tampered regions, and designed three functional modules to utilized the high-frequency features. Inspired by fine-grained classification, Zhou et al. [13] decomposed attention of single attention module network into multiple regions for collecting local features for deepfake detection.

Most of previous works bent their efforts to discriminate artifacts in human perceptible ways, leaving a bias of ignoring some critical invariance features intra-class. Therefore, instead of guiding feature extraction straight, we pay our attention to representation space and solve the problem through learning the invariant representations of natural faces and forgery faces both, which bring the intra-class representations close and push the inter-class representations apart, and influence feature extraction procedure indirectly.

**Contrastive Representation Learning.** Similar to the human cognitive processes, contrastive learning methods learn from multi-views of an identical object to keep the consistence of views from the same object in the representation space and punish the views sharing a high degree of similarity which belong to different objects. Intuitively, how to describe objects effectiveness through views which control the information captured by the representation is crucial to contrastive learning.

Several methods have been proposed to tackle the problem. Inspired by mixup augmentation, Wang et al. [34] added a static frame to every other frame to construct a distracting video sample which avoids the background cheating problem in the action recognition field. Instead of using Mixup as a data augmentation strategy, Zhu et al. [27] used feature transformation to extrapolate positive features and interpolate negative features to transform harder positives and negatives. As development of minimal sufficient statistics [30] and the Information Bottleneck theory [31, 32], Tian et al. [29] developed InfoMin principle to complete the InfoMax principle, which used the lower bound of NCE to formally describe a set of good views which should share the minimal information while keeping the maximal task-relevance information.

Limited to the setup of unsupervised learning, the latent relationships within data and how to utilize those relationships to guide a more effective representation learning are rarely discussed because of lacking label information. In addition, the improvement of positives construction strategy remains staying at the sample level which influences the network updates indirectly and does not independent from human intuition. Therefore, we take the procedure of deepfake synthesis into consideration and propose a novel forgery semantical guided pairing strategy to mine latent generation-related features, and design a hard feature mining and synthesizing method to simulate the potential marginal features.

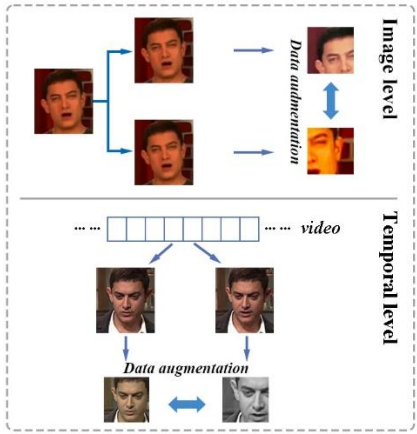

Figure 3: Our proposed image-level and temporal-level positives pairing strategy.

## 3. Proposed method

### 3.1. Overview

Instead of defining and representing artifacts at specific domain, we optimize the feature extraction procedure at the representation space through constraining intra-class consistence and inter-class inconsistence to bring the intra-class representations close and push the inter-class representations apart. Therefore, as illustrated in Figure 2, we propose a novel real-centric consistency learning method for deepfake detection which has been the first to introduce contrastive learning into deepfake detection field as far as we know. Specifically, we constraint feature extraction from two aspects, namely sample level and feature level. At the sample level, to mine latent generation-related features, we take the procedure of deepfake synthesis into consideration and propose a novel forgery semantical guided pairing strategy (see Sec. 3.3). At the feature level, based on the center of the natural face at the representation space, we design a hard positive mining and synthesizing method to simulate the potential marginal features. In addition, we design a hard negative fusion method, as an auxiliary, where the supervised contrastive loss is developed with margin to improve the discrimination of marginal features (see Sec. 3.4).

### 3.2. Preliminaries

We begin by briefly revisiting contrastive representation learning. In a typical contrastive learning method, which aims to bring the intra-class representations close and push the inter-class representations apart under self-supervision, two views are constructed as a pair of positive through applying stochastic data augmentation twice to an identical image. After that, an encoder and a project head are used to extract features and map these features into hypersphere. The contrastive loss is minimized to keep the consistence of views from the same image in the representation space and punish the views sharing a high degree of similarity which belong to other images. The InfoNCE loss [35] which maximizes the lower bound of mutual information between positives is commonly used as the contrastive loss. Inspired by soft-nearest neighbors loss [26], Khosla et al. [23] extended the self-supervised InfoNCE loss to the supervised setting which constructs positives in the class-level but still treats views belonging to the same class as input positive pairs.

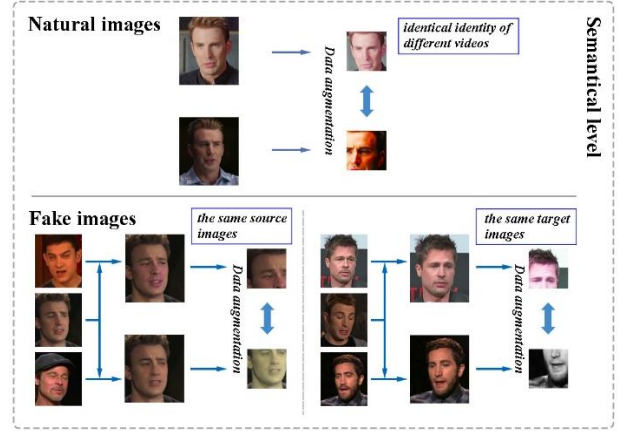

Figure 4: The proposed semantical-level positives pairing strategy.

### 3.3. Forgery semantical guided pairing strategy

During the forgery face generation procedure, generative methods make lifelike deepfakes from two existing videos of different identifications. They usually extract facial features of target videos and fuse them into faces regions of source videos to synthesis forgery faces [36, 37], and make the forgery faces adaptive to source images through blending methods. It seems that the background of manipulated videos remains the same as in the source videos, while the facial regions have been manipulated which contain target facial features and source facial features both. It reminds us that we could utilize the latent relationships within synthesis data and the temporal relationship to describe natural faces and forgery faces in a facial semantical irrelevant way and guide a more effectiveness representation learning.

Therefore, we propose a forgery semantical guided pairing strategy at the sample level. The positives are paired in three aspects: (1) image-level, (2) temporal-level, and (3) semantical-level. At the image-level, as shown in Figure 3, the same as regular contrastive learning, we treat two views of a single image as a pair of positive. Then, considered that the temporal relationship could provide variant descriptions of identical identity compared to views augmented from the same image and make a time-independent description. Utilizing the temporal relation-ship, we construct positives from two continuous frames at the temporal-level. At the semantical-level, the forgery faces synthesized from the same source images which contain the likely background texture could describe the fake images in a facial semantical irrelevant way, and the one synthesized from the same target images which contain the same facial features could describe the fake images in a background texture irrelevant way. As shown in Figure 4, the images synthesized from the same source images or the same target images are paired as positives, and the natural images in distinct videos while belonging to identical character are paired as positives similarly for the same reason. Besides, the stochastic data augmentation [25] is applied to each image to decrease the similarity between positive pairs further.

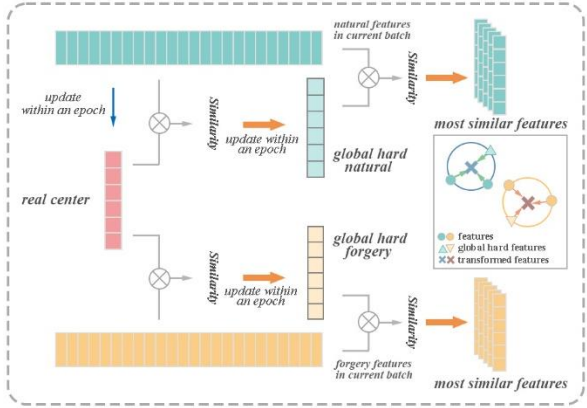

Figure 5: The hard positive transformation method simulates the potential marginal positives through transforming several positives.

| Methods | Celeb-DF-v2 | | | FF++ |
|---|---|---|---|---|
| | raw | c23 | c40 | |
| Xception [2] | 0.943 | 0.939 | 0.980 | 0.990 |
| Two-stream [14] | 0.965 | 0.964 | 0.990 | 0.994 |
| Our baseline [23] | 0.948 | 0.947 | 0.938 | 0.993 |

Table 1: Each model was trained on the Celeb-DF-v2 training set. The experimental results show that regardless of the compression ratio or data source, existing deepfake detection methods or even baseline methods perform well to the natural face data.

As demonstrated in the InfoMin principle [29], a good pair of positive should share minimal information while keep the maximal task-relevant information. Intuitively, the best positives here are those natural images of different identities or those fake images of distinct source faces and target faces. However, our experimental results reveal that excessive discrepancy between positive will underperform the representation learning, and details will be discussed in the Sec. 4.3.1.

### 3.4. Real-centric hard feature fusion method

The improvement of positives pairing strategy remains staying at the sample level which influences the network updates indirectly and does not independent from human intuition. In this section, we make constraints at the feature level, narrowing the intra-class discrepancy and widening the inter-class gap through simulating the potential marginal features.

As shown in Table 1, we discover from experimental results that regardless of the compression ratio or the data source, existing detection methods or even baseline methods are well performed on the natural data. It inspires us that the features extracted from natural faces are robust to the disturbance of comparison and the distribution shift is imperceptible between natural faces in distinct sources. We hypothesis that the features of natural faces sharing a vector space extended by a set of basic vectors, and there exist and only exist single center of natural faces. Therefore, the deepfake detection problem is converted into delimiting precise boundary of natural faces at the representation space, and we propose a real-centric hard feature fusion method which consists of 4 components: real-centric hard samples mining, hard positives transformation, hard negative mixing, and the developed supervised contrastive margin loss.

**Real-centric hard samples mining.** Supposing $z$ denotes the normalized features at the hypersphere, and $z_{real}$ is the natural face features, we use $N$ to denote the number of features, and $N_{real}$ is the number of natural face features. The real center $c$ is calculated as follows:

$$c = \frac{1}{N_{real}} \sum_{n=0}^{N_{real}} z_{real}^n \quad (1)$$

Here, we utilize L2 normalization and update the center within an epoch, and the $N_{real}$ is equal to the number of natural faces which have shown in the current epoch.

Based on the real center, as illustrated in Figure. 5, we mine the marginal features as hard samples. Specifically, we use similarity to measure the discrepancy of the center and features. With the goal of delimiting precise boundary of natural faces, the marginal natural features and the fake features which share most similarity with natural features are more important for us. Therefore, the least similar natural features and the most similar fake features with the center are seen as hard features in our proposed method.

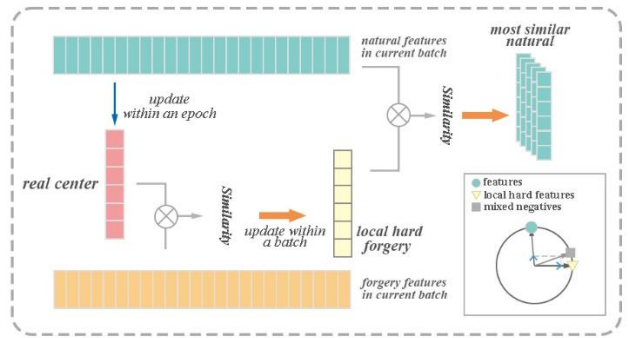

Figure 6: The hard negative mixing method simulates the potential marginal negatives through mixing local hard forgery feature and its most similar natural feature.

Formally, let $Z_{real} = \{z_{real}\}$, $Z_{fake} = \{z_{fake}\}$ denote the representation sets of natural faces and the forgery faces respectively. Calculating the similarity of the center and each feature to acquire the similarity sets of natural faces and forgery faces $S_{real} = \{s_{real}\}$, $S_{fake} = \{s_{fake}\}$, we select the corresponding representations of the top-k most similar features from the fake representation set $Z_{fake}$ and the corresponding representations of the tail-k most similar features from the natural representation set $Z_{real}$ to compose the hard fake set $Z_{top} = \{z_{top}\}$ and the hard natural set $Z_{tail} = \{z_{tail}\}$. We update those hard representation sets within an epoch which is the same with the real center to provide a global description of marginal features as well as averting the impact of the imprecise historical marginal features. In addition, we keep a local description of hard fake representations through updating hard fake representations within a batch either.

**Hard positives transformation.** To narrow the intra-class discrepancy, we simulate the potential marginal positives through transforming several positives.

For each global hard features $z_{hg}$ in the hard natural set $Z_{tail} = \{z_{tail}\}$ and hard fake set $Z_{top} = \{z_{top}\}$, we find their $M$ most similar local features through calculating their similarity with each intra-class features in the current batch and selecting the $M$ most similar features.

Formally, for each global hard feature $z_{hg}$ and their $M$ most similar local features $z_{mbr}$, the transformed positives $z_{tp}$ are calculated as follows:

$$z_{tp} = \frac{1}{M+1}\left(z_{hg} + \sum_{m=0}^{M} z_{mbr}^m\right) \quad (2)$$

Then we use L2 normalization. Due to that the global hard features are existing marginal features which have shown in the current epoch and their corresponding $m$ local features are real existing features in the current batch, the hard positives we simulated could reflect the real distribution of marginal features, and the combination of global hard features and their local similarity features leads the diversity of synthesized features and promotes the representation learning procedure.

**Hard negatives mixing.** We design a hard negatives mixing strategy, which is shown in Figure. 6, to clearly depict the boundary of natural faces at the feature level.

To each local hard fake feature $z_{hl}$, we calculate their similarity with each natural feature in the current batch, and select the $V$ most similar natural feature $z_{vbr}$. The mixed hard negatives are calculated as follows:

$$z_{mn} = \lambda \cdot z_{hl} + (1-\lambda) \cdot z_{vbr} \quad (3)$$

Where $\lambda \sim Beta(0.8, 0.8)$ is a coefficient. Beta distribution is parameterized by two positive shape parameters, denoted by $\alpha$ and $\beta$, that appear as exponents of the random variable and control the shape of the distribution. We set $\alpha = \beta < 1$ to make the probability density function in a U-shape, and restrict an extreme condition of the marginal negative features. Then L2 normalization is used to map the $z_{mn}$ into hypersphere.

**Supervised contrastive margin loss.** We develop a supervised contrastive margin loss to take advantage of the mixed negatives and widen the inter-class gap, defined as follows:

$$\mathcal{L} = \sum_{i \in I} \frac{-1}{|P|} \sum_{j \in P'} \log \frac{\exp(Sim(z_i, z_j)/\tau)}{\sum_{a \in A'} \exp(Sim(z_i, z_a)/\tau) + Margin} \quad (4)$$

where

$$Margin = \sum_{v \in A_n} \exp(Sim(z_i, z_v)/\tau) \quad (5)$$

Where $i \in I \equiv \{1, ..., 2N\}$, $N$ is the batch size, $a \in A \equiv I \setminus \{i\}$, $P$ indicates the index set of all positives in the batch distinct from $i$. Keeping $u$ global hard naturals and global hard fakes respectively, then $A_p = \{1, ..., 2u\}$ is the index of transformed positives set, and $A_n = \{1, ..., sM\}$ is the mixed negatives set where $s$ is the number of local hard forgery. Then extending $P$ and $A$ by transformed positives, we have $P' \equiv P \cup A_p$ and $A' \equiv A \cup A_p$. Let $y$ denotes the label, and $j = P_i \equiv \{j \in A, if\ y_j = y_i\}$. $\tau$ is the temperature scale.

## 4. Experiments

In this section, we introduce the overall experiment settings firstly and evaluate the efficiency of our proposed method. Then, ablation study is performed to analyze the effects of each strategy in our methods.

### 4.1. Experiment setting

**Dataset** In order to take advantage of the data semantical correlations to guide the positive pairing, our experiments are mainly conducted on Celeb-DF-v2 [1] dataset. Celeb-DF-v2 is a newly proposed large-scale dataset with high visual quality containing 890 real videos and 5639 fake videos. The real videos includes 590 celebrity videos from 61 celebrities and 300 additional videos. The fake videos are synthesized through cross manipulating to those celebrities.

To simulate the internet interference environment, we process the dataset into light compression version by setting the quantization parameter equal to 23 (c23) and heavy compression version by using a quantization of 40 (c40).

| Methods | raw | | | | c23 | | | | c40 | | | |
|---|---|---|---|---|---|---|---|---|---|---|---|---|
| | TPR | FPR | AUC | ACC | TPR | FPR | AUC | ACC | TPR | FPR | AUC | ACC |
| Xception [2] | 94.34 | 98.01 | 99.45 | 97.12 | 93.87 | 96.07 | 99.03 | 95.68 | 98.04 | 16.17 | 78.99 | 44.10 |
| Two-stream [14] | 96.52 | 99.61 | 99.87 | 98.51 | 96.37 | 97.74 | 99.62 | 97.27 | **99.04** | 12.20 | **82.38** | 41.77 |
| Multi-attention [13] | **97.12** | 99.73 | **99.88** | **98.81** | **97.15** | 97.38 | 99.54 | **97.30** | 98.67 | 14.86 | 80.61 | 43.40 |
| Sup. Con [23] | 94.75 | 99.50 | 99.79 | 97.82 | 94.73 | 98.16 | 99.56 | 96.99 | 93.84 | 24.13 | 73.03 | 47.87 |
| Ours. | 93.77 | **99.80** | 99.86 | 97.67 | 93.06 | **99.18** | **99.66** | 97.09 | 97.09 | 27.42 | 78.44 | **51.27** |

Table 2: Compared with previous works on the three quality versions of Celeb-DF-v2 dataset, our proposed method shows outstanding comprehensive performance on each version.

**Implementation details** All run of our framework are based on a symmetry architecture where the two streams of encoders and projectors share parameters. Besides, we use Xception [16] without the last FC layer pretrained on the ImageNet [20] as encoder and a two layers perceptron as projector which projects 2048-d vectors into 128-d vectors for the proposed method. In the training stage, we first train an encoder and a projector with supervised contrastive margin loss for 100 epochs, then the projector is removed and the encoder is frozen to train the linear classifier with cross entropy loss for 10 epochs. In the testing stage, we just use the encoder and the classifier to extract features and map the encoded features into logits.

The networks are optimized by SGD. We set the base learning rate as 0.01. The momentum is set as 0.9. The batch size is set as 128. The temperature is set as 0.1. The $u$ is set as 32 and $s$ is set as 4. The $k$ is set as 2, and $M$ is set as 4.

Before training, we split each video into frames and extract face regions in advance. We adopt dlib [33] to crop the face and apply a conservative crop to enlarge the face region. To avoid the impact of heavily class imbalance in the training set, we under-sample the fake data and let the number of each class equally.

### 4.2. Experimental results

To verify the performance of the proposed method, we compare our approach with existing methods on Celeb-DF-v2 dataset and its low-quality versions. All models are trained on the original version of the training set and tested on the three quality versions of the testing set. We use true positive rate (TPR), false positive rate (FPR), area under curve (AUC) and accuracy (ACC) as metrics to measure the overall performance of each method. The results are listed on Table 2.

The experimental results indicate that original supervised contrastive learning framework has shown the robustness against compression, while it shows significant shortcoming at the low quality natural faces. When our proposed pairing strategy and hard feature fusion method are applied, although the TPR of the original version is slightly underperformed, its outstanding performance at the low quality testing set and excellent overall quality demonstrate the effectiveness and the robustness of our proposed method.

### 4.3. Ablation study

#### 4.3.1 Forgery semantical guided pairing strategies

In this section, we will show the relevant results of the positive pairing strategy experiments in detail to validate the strategy. Considering the relevance implicated in the data, we design four pairing strategies and conduct experiments under the same data augmentation operations:

**Instance-level pairing** As in regular contrastive learning, each image is augmented twice to acquire the positive pair.

**Temporal-level pairing** Taking temporal relationship into consideration, aside from instance-level pairing, the continuous two frames in a video are paired as positives.

**Forgery semantical guided pairing** As mentioned in Sec 3.3, the implicated relationships of data are described in three aspects: image level, temporal level, and semantical level.

**Class-level pairing** In view of all potential relationships within the data, in addition to the forgery semantical guided pairing followed positives, the fake images synthesized from distinct characters or the natural images belonging to distinct identities are paired.

It is noteworthy that supervised contrastive loss treats the input pairs and the images in the same class with inputs in the current batch as positives either. Hence, the class-level pairs show a situation of the forgery semantical guided pairs that more pairs sharing with minimal information while keeping the maximal task-relevant information are contained in the training data.

As shown in Table 3, due to that the adjunction of temporal relevance, the network outperforms other strategies on the testing set except for the low quality version (c40). We suggest this is due to that the temporal pairs provide a more slightly diversified description of data than the instance-level pairs, which contributes to representation learning, while that diversity is insufficient for learning the intra-class consistency. Although the pairs which class-level pairing strategy products seem to follow the InfoMIN principle [29], the experimental results demonstrate that because of the absence of explicit guidance, the model underperforms on the natural face, and it is nonnegligible that the robustness against the compression exhibits in the experiments on several low quality data.

| Methods | raw | | | c23 | | | c40 | | |
|---|---|---|---|---|---|---|---|---|---|
| | TPR | FPR | ACC | TPR | FPR | ACC | TPR | FPR | ACC |
| Instance-level | 94.8 | 99.5 | 97.8 | 94.7 | 98.1 | 97.0 | 93.8 | 24.1 | 47.9 |
| Temporal-level | **96.4** | 99.5 | **98.4** | **96.6** | 98.0 | **97.5** | **97.1** | 17.4 | 44.6 |
| Forgery semantical | 94.2 | 99.6 | 97.9 | 94.5 | 98.6 | 97.2 | 95.0 | 27.5 | 50.5 |
| Class-level | 90.0 | **99.8** | 96.3 | 85.7 | **99.5** | 94.8 | 88.4 | **41.6** | **57.6** |

Table 3: Experimental results of different pairing strategies.

| Methods | raw | | | c23 | | | c40 | | |
|---|---|---|---|---|---|---|---|---|---|
| | TPR | FPR | ACC | TPR | FPR | ACC | TPR | FPR | ACC |
| Linear | 92.7 | 99.6 | 97.2 | **91.7** | 99.0 | 96.5 | 96.2 | 28.8 | 51.7 |
| Linear + smooth | 92.8 | 99.3 | 96.9 | 91.6 | 98.4 | 96.0 | 92.1 | **36.4** | **55.3** |
| Smooth | **93.0** | **99.9** | **97.4** | 90.9 | **99.4** | **96.5** | **97.0** | 26.8 | 50.7 |

Table 4: Experimental results of different feature fusion methods.

| Methods | raw | | | c23 | | | c40 | | |
|---|---|---|---|---|---|---|---|---|---|
| | TPR | FPR | ACC | TPR | FPR | ACC | TPR | FPR | ACC |
| Sup. Con [23] | **94.8** | 99.5 | **97.8** | **94.7** | 98.2 | **97.0** | 93.8 | 24.1 | 47.9 |
| + 32 pos. | 93.3 | 99.9 | 97.6 | 91.3 | **99.6** | 96.8 | 98.7 | 23.7 | 49.2 |
| + 32 pos., 16 neg. | 93.0 | 99.9 | 97.4 | 90.9 | 99.4 | 96.5 | 97.0 | 26.8 | 50.7 |
| + 32 pos., 32 neg. | 92.0 | **99.9** | 97.1 | 90.8 | 99.5 | 96.5 | **97.1** | **30.0** | **52.9** |

Table 5: Experimental results of different quantity of additive fused features.

| Methods | raw | | | c23 | | | c40 | | |
|---|---|---|---|---|---|---|---|---|---|
| | TPR | FPR | ACC | TPR | FPR | ACC | TPR | FPR | ACC |
| Sup. Con [23] | **94.8** | 99.5 | **97.8** | **94.7** | 98.2 | **97.0** | 93.8 | 24.1 | 47.9 |
| As fakes | 91.0 | 99.8 | 96.7 | 91.3 | 97.3 | 95.3 | **99.1** | 19.8 | 46.8 |
| As negatives | 93.0 | **99.9** | 97.4 | 90.9 | **99.4** | 96.5 | 97.0 | **26.8** | **50.7** |

Table 6: Experimental results of different positions of fused feature fusion methods.

#### 4.3.2 Real-centric hard feature fusion methods

**How should we fuse positives?** Several previous works fused features between negatives through linear combination [27, 28]. We borrow this idea as the control group and conduct comparative experiments. As listed in Table 4, "Linear" indicates that we combine 8 hard features and their top 4 most similar features through linear combination. "Linear + smooth" denotes that after the features have been combined, we select top 2 most similar features in current batch for each fused features and smooth those fused features through Eq. 3 and Eq. 4. "Smooth" means we fuse those features as described in Sec. 3.4 where 32 hard features are selected to be fused with their top 2 most similar features. The results demonstrate that the smooth strategy makes the synthesized features conform to the real data distribution which promotes the representation learning.

**How many features should be fused?** The number of negatives could influence the contrastive loss directive, and overfull extra features could lead to a rapidly variety of the gradient which is harmful to the performance of detection. Therefore, we conduct comparative experiments to discriminate the influence and the effectiveness of the fused features. As shown in Table 5, the experimental results illustrate that both the fused positives and the fused negatives could improve the performance of robustness against the compression while slightly decreasing the precision of detection for the natural faces. Under discretionary trade-offs, a size of 32 additional transformed positives and 16 additional mixed negatives strike a good balance between different quality versions.

**Where should we utilize fused negatives?** When we aim at delimiting precise boundary of natural faces at the representation space, it makes sense that the inter-class fusion features can be treated as fake image features. However, as shown in Table 6, it is harmful to the overall performance of model that we roughly take the inter-class fusion features as fake features which disturb the representation learning of fake representations. By contrast, it is a more suitable choice to use the supervised contrastive margin loss and see inter-class fusion features.

## 5. Conclusions

In this paper, we have proposed a novel real-centric consistency representation learning method for deepfake detection through learning the invariant representations of both classes instead of representing artifacts at a specific domain. We constrain feature extraction from two aspects, namely sample level and feature level. At the sample level, we propose a novel forgery semantical guided pairing strategy to mine latent generation-related features. In addition, based on the centers of natural faces at the representation space, we design a hard positive synthesizing method to simulate the potential marginal features and a hard negative fusion method to improve the discrimination of marginal features at the feature level. The comprehensive experiments demonstrate the effectiveness of our proposed method.

Despite our method exhibiting robustness of compression, there still remains the room of improvement for the real-world application. In addition, the generalization to unseen generative methods is a more challenging research direction which has not been deeply discussed in this paper. Therefore, our future research will focus on how to represent features in a more effective and generative way.